\documentclass{article}

\usepackage[preprint]{neurips_2026}

\usepackage{enumitem}
\usepackage{amsmath}
\usepackage{wrapfig}

\usepackage[utf8]{inputenc} 
\usepackage[T1]{fontenc}    
\usepackage{hyperref}       
\usepackage{url}            
\usepackage{booktabs}       
\usepackage{amsfonts}       
\usepackage{nicefrac}       
\usepackage{microtype}      
\usepackage[table]{xcolor}  
\usepackage{graphicx}       
\usepackage{multirow}       
\usepackage[ruled,vlined]{algorithm2e}
\usepackage{graphicx}
\usepackage{subcaption}
\title{AGC: Adaptive Geodesic Correction for Adversarial Robustness on Vision-Language Models}

\author{
  Zhiwei Li\textsuperscript{1,3}\thanks{Equal contribution.}
  \quad Jiacheng Xue\textsuperscript{2}\footnotemark[1]
  \quad Weining Wang\textsuperscript{3}
  \quad Ajian Liu\textsuperscript{3} \\
  \textbf{Xingyu Gao}\textsuperscript{3}
  \quad \textbf{Zhenan Sun}\textsuperscript{1,3}
  \quad \textbf{Qi Li}\textsuperscript{1,3}\thanks{Corresponding author.} \\
  \textsuperscript{1}NLPR \& MAIS, Institute of Automation, Chinese Academy of Sciences \\
  \textsuperscript{2}School of Computer Science and Engineering, Central South University \\
  \textsuperscript{3}University of Chinese Academy of Sciences \\
  {\small
lizhiwei2023@ia.ac.cn, 8208230705@csu.edu.cn, weining.wang@nlpr.ia.ac.cn,}\\
{\small ajian.liu@ia.ac.cn, gaoxingyu@ime.ac.cn, \{znsun, qli\}@nlpr.ia.ac.cn
}
}

\begin{document}

\maketitle

\begin{abstract}
Vision-language models like CLIP have demonstrated remarkable zero-shot transfer capabilities. However, their susceptibility to imperceptible adversarial perturbations remains a critical security concern. While test-time defenses offer a pragmatic solution for deployed models, existing approaches typically rely on gradient-based optimization during inference, incurring significant computational overhead. In this paper, we revisit the role of data augmentation in CLIP robustness and observe that augmentations are not equally effective: specific augmentations consistently provide robust geometric cues that align with correct class semantics in the hyperspherical feature space. Based on this, we propose Adaptive Geodesic Correction (AGC), a training-free defense mechanism that requires no parameter updates. AGC identifies a reliable augmentation as a geometric anchor and corrects the input feature towards it, utilizing an adaptive step size to balance robustness against clean accuracy preservation. AGC achieves superior performance across eight fine-grained datasets and three CLIP backbones, improving average robust accuracy by 44.4\% over state-of-the-art baseline while delivering a 10$\times$ reduction in inference latency. Our findings reveal a fundamental geometric property of CLIP features, offering a highly efficient and effective paradigm for robust multimodal deployment. \url{https://github.com/lizhiwei23/AGC}

\end{abstract}

\section{Introduction}

Vision-language models such as CLIP~\cite{radford2021learning} have achieved remarkable success in multimodal understanding and open-vocabulary recognition by learning a shared image-text embedding space~\cite{jia2021scaling, chenvlp, wang2023large, zhang2024vision}. Their strong zero-shot transfer has led to widespread real-world deployment of publicly available pre-trained checkpoints. However, prior studies have shown that CLIP is vulnerable to imperceptible adversarial perturbations~\cite{mao2022understanding,li2024one,zhou2024few,schlarmann2024robust}, motivating recent test-time defense methods~\cite{sheng2025r,xing2025clip,li2025ttp,wang2025tapt,tong2025zero}, which can cause completely incorrect predictions. Since practitioners usually adopt off-the-shelf models without retraining, improving CLIP’s adversarial robustness at test time is of particular practical importance.

Existing CLIP defenses fall into training-time and test-time approaches. Training-time methods, such as adversarial fine-tuning and adversarial prompt learning~\cite{mao2022understanding, li2024one, zhou2024few, schlarmann2024robust}, improve robustness by optimizing model parameters on labeled data and adversarial examples. However, they require extra supervision and substantial training cost, and often generalize poorly to unseen classes, datasets, or backbones. 
\begin{wrapfigure}{r}{0.55\textwidth}
    \centering
    \includegraphics[width=\linewidth]{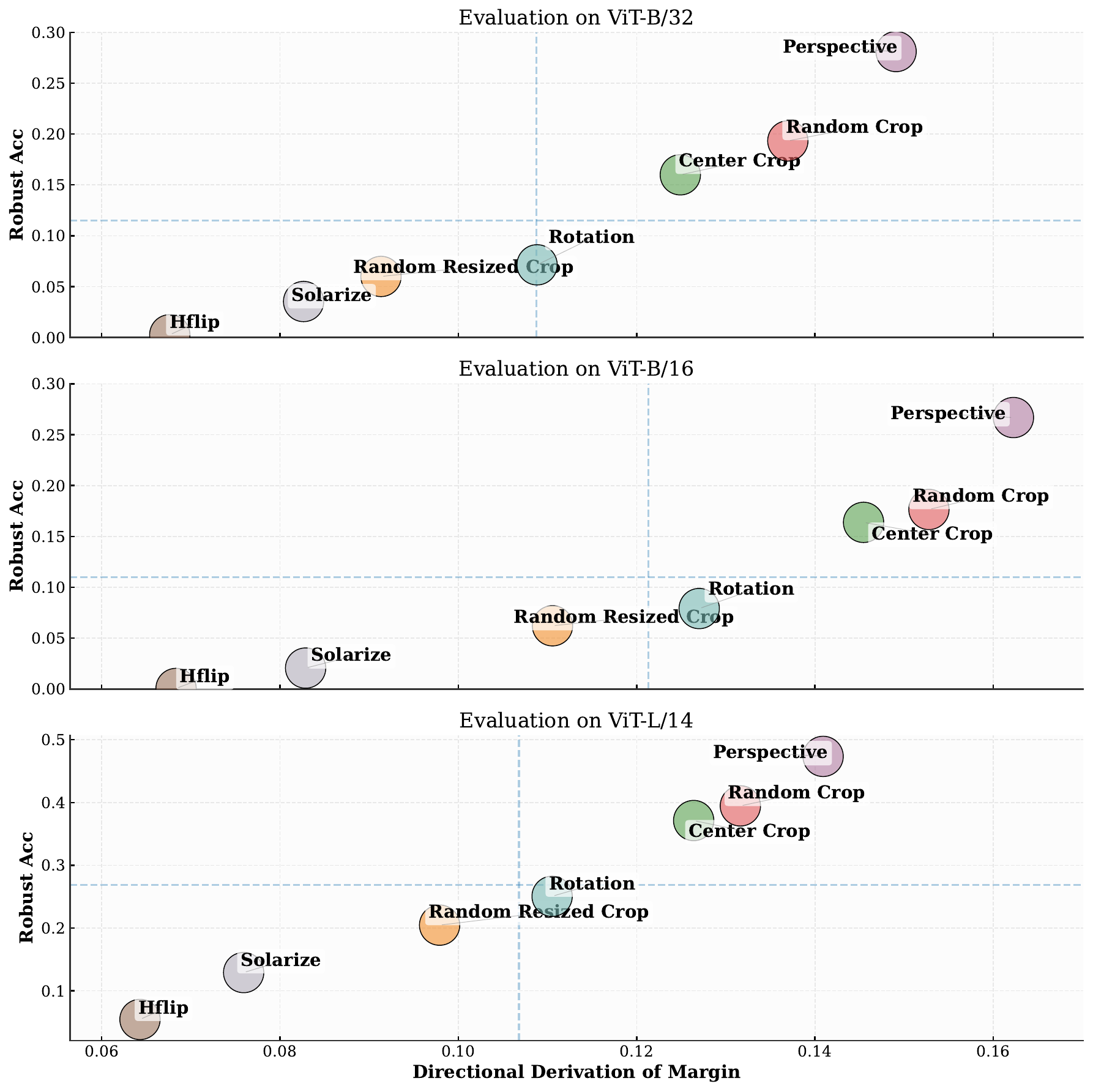}
    \caption{Relationship between first-order derivation and robustness using different augmentations across three CLIP backbones.}
    \label{fig:three_scatter}
\end{wrapfigure}
Test-time defenses are more suitable for deployment, but most are not truly training-free: they still perform backpropagation during inference to update prompts or other learnable components, which incurs considerable overhead~\cite{shu2022test,feng2023diverse,abdul2023align,sheng2025r,li2025ttp,wang2025tapt}. In addition, although most of these methods use augmented views, they mainly treat them as inputs for ensembling or adaptation, without asking why some augmentations are consistently more robust than others. As a result, they often use many augmentations indiscriminately, further slowing inference.

In this paper, we revisit the role of augmentation in CLIP robustness and propose Adaptive Geodesic Correction (AGC), an efficient test-time defense with no parameter updates. Our key observation is that augmentations are not equally robust: some consistently yield much stronger adversarial robustness than others, and can even surpass the strongest existing test-time defenses. This pattern is stable across datasets and CLIP backbones, suggesting that the robustness gains from augmentation are highly structured rather than incidental. To understand why certain augmentations are consistently more effective, we analyze the relationship between the original adversarial feature and its augmented counterparts in CLIP's normalized feature space. Since CLIP classifies by cosine similarity between normalized image and text embeddings, image features lie on a unit hypersphere, where geodesics naturally characterize feature directions. Because adversarial perturbations can be viewed as pushing features away from the correct semantics, a natural question is whether the high robustness of certain augmentations stems from their ability to pull adversarial features back toward the correct semantics.

To answer this question, we first quantify correct semantics through the classification margin, defined as the similarity to the ground-truth text feature minus the largest similarity to any incorrect class. For each augmentation, we measure the first-order change of this margin along the geodesic from the original adversarial feature to the augmented feature. Empirically, we observe a strong positive correlation between this first-order margin improvement and the robustness (Fig.~\ref{fig:three_scatter}). This result supports a geometric interpretation of augmentation robustness: effective augmentations move adversarial features along directions that are more aligned with correct semantics, and therefore have the potential to serve as reliable anchors for feature correction.

Motivated by this geometric insight, AGC selects the most reliable augmentation as an anchor and uses it to define a correction direction for the original feature. A central challenge is that test-time inputs may be either clean or adversarial: clean samples should be preserved, whereas adversarial examples require stronger correction. Since this distinction is unknown at inference time, AGC adopts an adaptive step size based on two signals: the deviation between the original feature and the anchor, and the consistency among augmented features. When both signals are strong, AGC applies a larger correction, suggesting that the feature has drifted away from the clean semantics; otherwise, it performs a conservative update to avoid disturbing clean semantics.

AGC uses only a small set of informative augmentations and requires no gradient-based optimization, substantially reducing inference-time overhead. Across eight fine-grained classification datasets and three CLIP backbones, AGC achieves an average robust accuracy of 89.2\%, significantly outperforming the strongest baseline test-time defense (44.8\%), while maintaining clean accuracy comparable to the vanilla CLIP. It also delivers more than 10$\times$ faster inference than the existing SOTA test-time defense, highlighting its practical value for real-world deployment. In summary, our contributions are as follows:

\begin{itemize}[leftmargin=*]
    \item We revisit the role of augmentation in CLIP robustness and explain why different augmentations exhibit markedly different effectiveness: stronger augmentations tend to steer adversarial features along geodesic directions that are better aligned with the correct class semantics.
    \item We propose Adaptive Geodesic Correction (AGC), a truly training-free defense for CLIP. AGC selects a reliable augmentation as a geometric anchor, uses it to define a correction direction, and adaptively adjusts the input feature along this direction, without any parameter updates during inference.
    \item We demonstrate that AGC achieves an excellent balance between robustness, clean accuracy, and inference efficiency across eight fine-grained datasets and three CLIP backbones, substantially outperforming prior test-time defenses.
\end{itemize}

\section{Related Work}
\subsection{Adversarial Attack and Defense}
Adversarial attacks aim to deceive a model by injecting imperceptible perturbations into input samples, leading to severe prediction errors~\cite{szegedy2013intriguing}. Existing attacks can be broadly categorized into gradient-based attacks~\cite{MIM,BIM,PGD,FGSM}, transformation-based attacks~\cite{input_div, translation, bsr}, and feature-level attacks~\cite{FIA,NAA}. Among them, gradient-based attacks are the most representative and widely used in practice. In particular, Projected Gradient Descent (PGD)~\cite{PGD} has become the standard attack paradigm for evaluating adversarial defenses due to its strong empirical effectiveness and widespread adoption in prior work.

To improve robustness against such attacks, a large body of defense methods has been proposed, among which adversarial training and its variants remain the dominant paradigm~\cite{zhang2019theoretically, wu2020adversarial}. For vision-language models such as CLIP, most existing robustness methods also follow this training-time paradigm by incorporating adversarial examples into optimization. Representative approaches include adversarial fine-tuning methods such as TeCoA~\cite{mao2022understanding} and PMG-AFT~\cite{wang2024pre}, which improve the robustness of the CLIP image encoder and transfer it to downstream tasks, as well as adversarial prompt-learning methods such as APT~\cite{li2024one}, which optimize robust prompts while keeping most model parameters frozen. Although effective, these methods typically require labeled data, repeated adversarial example generation, and substantial training cost, and may compromise clean accuracy or zero-shot generalization under unseen distributions. 

\subsection{Test-time Adaptation for VLMs}
Test-time methods improve model behavior during inference without full retraining, making them attractive for practical CLIP deployment. Early methods in this line were mainly designed for clean distribution shifts rather than adversarial robustness. For example, TPT~\cite{shu2022test}, DiffTPT~\cite{feng2023diverse}, PromptAlign~\cite{abdul2023align}, and TPS~\cite{sui2025just} adapt prompts at test time to improve zero-shot generalization, while MTA~\cite{zanella2024test} enhances robustness to distribution shifts through training-free multi-view aggregation. However, these methods generally assume clean inputs and do not explicitly address adversarial perturbations.

More recent work has begun to study adversarial robustness at test time. R-TPT~\cite{sheng2025r} improves robustness by optimizing textual prompts with a pointwise entropy objective, while TTP~\cite{li2025ttp} adopts a detect-then-adapt pipeline that first identifies adversarial examples through padding-induced feature shifts and then performs robust inference with trainable padding and similarity-aware ensembling. TTC~\cite{xing2025clip} avoids updating model parameters, but still performs a PGD-based counterattack in the input space to break the false stability of adversarial examples. Although these methods are framed as test-time defenses, they all require backpropagation during inference, and therefore are not strictly training-free. This introduces nontrivial computational overhead and limits their practical efficiency.

\section{Methodology}

\subsection{Preliminaries}

\paragraph{CLIP zero-shot classification.}
CLIP comprises an image encoder $g(\cdot)$ and a text encoder $f(\cdot)$, which project images and texts into a shared embedding space.
For a $C$-class classification task with class labels $\{y_c\}_{c=1}^{C}$, each class name is inserted into a prompt template $T(\cdot)$, and the corresponding normalized text feature is defined as
\begin{equation}
t_c=\frac{f(T(y_c))}{\|f(T(y_c))\|}, 
\qquad c=1,\dots,C.
\end{equation}
Given an input image $x$, its normalized visual feature is
\begin{equation}
z=\frac{g(x)}{\|g(x)\|}.
\end{equation}
Zero-shot classification is performed by comparing cosine similarities between the visual feature and text features:
\begin{equation}
\hat{y}=\arg\max_{c} z^\top t_c .
\end{equation}

\paragraph{Hyperspherical geometry.}
Since CLIP features are $\ell_2$-normalized, both visual and textual embeddings lie on the unit hypersphere, denoted by
$\mathbb{S}^{d-1}=\{x\in\mathbb{R}^d:\|x\|_2=1\}$.
For two visual features $z,a\in\mathbb{S}^{d-1}$, their geodesic distance is characterized by the angle
\begin{equation}
\theta(z,a)=\arccos(z^\top a).
\end{equation}
The unit tangent direction from $z$ toward $a$ is obtained by projecting $a$ onto the tangent space at $z$:
\begin{equation}
u(z,a)=
\frac{a-(a^\top z)z}
{\|a-(a^\top z)z\|}.
\label{uza}
\end{equation}
Accordingly, the geodesic starting from $z$ toward $a$ is
\begin{equation}
\gamma(t)=\cos(t)z+\sin(t)u(z,a), \qquad t\in[0,\theta(z,a)].
\end{equation}
In particular, $\gamma(0)=z$ and $\gamma(\theta(z,a))=a$.
This geometric formulation underlies feature correction on the CLIP hypersphere. Note that, in our method, $t$ is not restricted to $[0,\theta(z,a)]$, since the selected augmented point $a$ serves only as an anchor rather than the final target. 
Instead, it acts as an intermediate reference that guides the adversarial feature along the geodesic toward a region better aligned with the correct semantics.

\begin{figure}[!t]
  \centering
  \includegraphics[width=\linewidth]{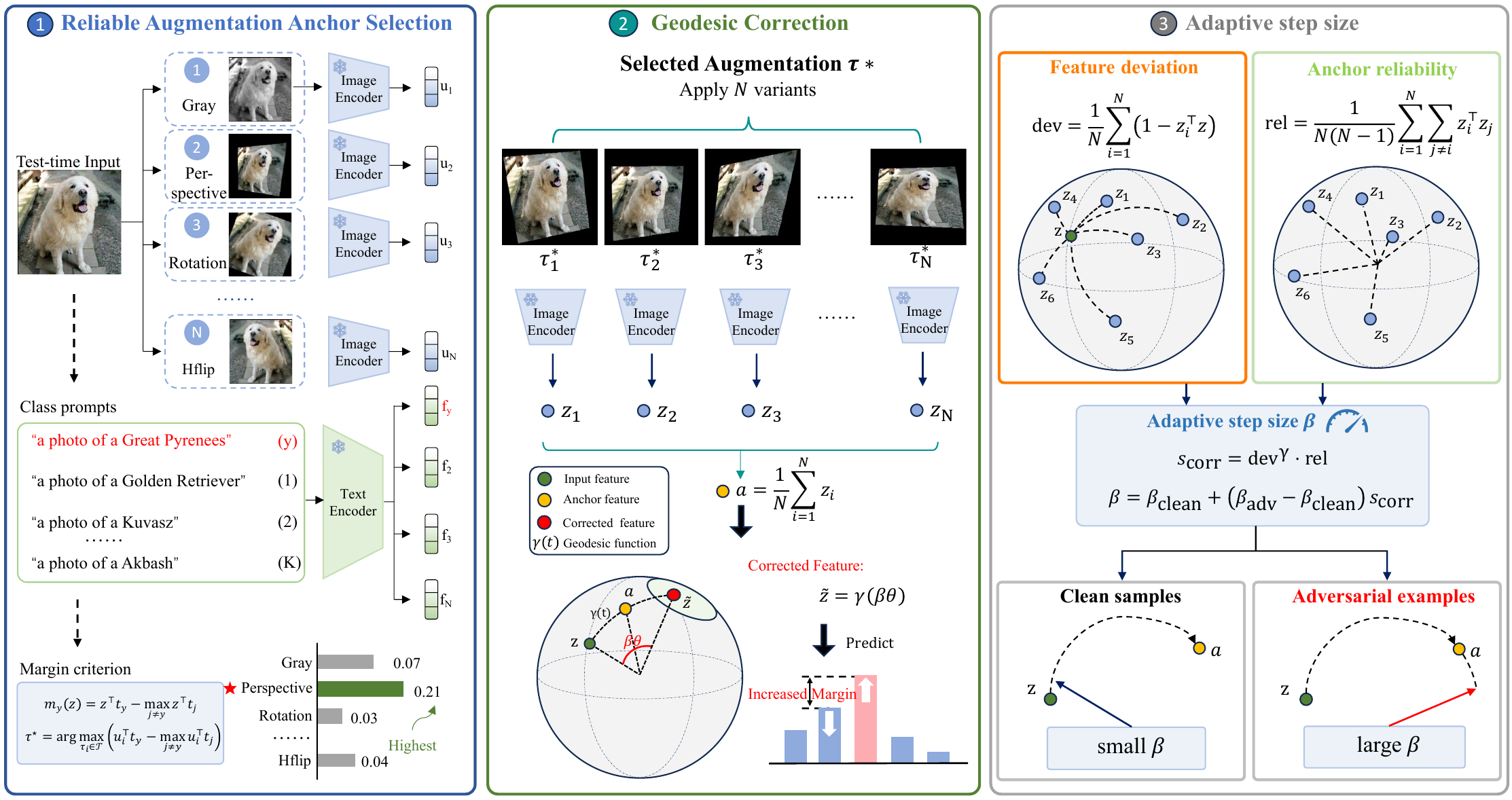}
  \caption{Overview of Adaptive Geodesic Correction (AGC). AGC first selects a reliable augmentation anchor using the margin criterion. The original feature is then corrected along the geodesic toward this anchor on CLIP's unit hypersphere. The correction step size is adaptively determined by feature deviation and anchor reliability, allowing AGC to improve adversarial robustness while preserving clean-sample semantics, without requiring backpropagation-based test-time updates.}
  \label{fig:model_arc}
\end{figure}

\subsection{Augmentation Evaluation by Classification Margin}

We next analyze why different augmentation types exhibit markedly different robustness in CLIP.
Our central hypothesis is that a robust augmentation should provide a reliable geometric anchor, namely, an augmented feature whose geodesic direction from the original feature is better aligned with the semantics of the ground-truth class.

Let $\mathcal{T}=\{\tau_1,\tau_2,\dots,\tau_K\}$ denote a set of candidate augmentation types.
Given an input image $x$ with normalized feature $z$,
for each augmentation type $\tau_i\in\mathcal{T}$, we compute the normalized augmented feature
\begin{equation}
a_i=\frac{g(\tau_i(x))}{\|g(\tau_i(x))\|},
\end{equation}
and the corresponding geodesic direction $u_i = u(z,a_i)$, which is the unit tangent vector at $z$ pointing toward $a_i$ computed in Eq.~\ref{uza}.

To quantify whether this direction is semantically beneficial, we consider the classification margin on the CLIP hypersphere. For class label $y$, the margin of $z$ is defined as
\begin{equation}
m_y(z)= z^\top t_y-\max_{j\neq y} \{z^\top t_j\},
\end{equation}
where $t_y$ is the text feature of the ground-truth class.
A larger margin indicates stronger alignment with the correct class and greater separation from competing classes.

Its directional derivative along $u_i$ is
\begin{equation}
\nabla_{u_i} m_y(z)
=
u_i^\top t_y
-
u_i^\top t_{j^*},
\qquad
j^*=\arg\max_{j\neq y} z^\top t_j.
\end{equation}

which measures the first-order change in the margin when moving from $z$ along the geodesic direction induced by augmentation $\tau_i$. Based on this margin score, we evaluate each augmentation type. For fairness, for every augmentation involving stochastic transformation strength, we consider three levels of intensity, weak, medium, and strong, and average their scores, so as to obtain a more balanced assessment of each augmentation.

As shown in Fig.~\ref{fig:three_scatter}, this margin-based score is strongly correlated with the adversarial robustness achieved by different augmentation types across datasets and backbones. This finding reveals the potential of such augmentations to serve as reliable anchors for correcting adversarial features, as the geodesic direction from the original adversarial feature to the augmented view provides a good approximation of the clean semantic direction. Among all candidates, \texttt{RandomPerspective} consistently yields the largest scores. We therefore use \texttt{RandomPerspective} as the default augmentation type in AGC. Importantly, this margin-based score is used only for analysis and augmentation evaluation, and does not enter the test-time inference procedure.

\subsection{Adaptive Geodesic Correction}

After selecting the augmentation type $\tau^\star$, AGC uses it to construct a reliable geometric anchor for feature correction.
As shown in the previous section, $\tau^\star$ is the augmentation whose induced geodesic direction is most favorable for increasing the classification margin.
To reduce the variance of a single augmented view, we generate $N$ stochastic views under $\tau^\star$ and aggregate their features to form the anchor.

Specifically, for the input image $x$, we compute
\begin{equation}
z_i=\frac{g(\tau_i^\star(x))}{\|g(\tau_i^\star(x))\|},
\qquad i=1,\dots,N,
\end{equation}
where each $\tau_i^\star$ denotes an independent stochastic instantiation of augmentation type $\tau^\star$.
The anchor feature is then defined as
\begin{equation}
a^\star=\frac{\sum_{i=1}^{N} z_i}{\left\|\sum_{i=1}^{N} z_i\right\|}.
\end{equation}
Importantly, $a^\star$ is not treated as the final destination of correction.
Instead, it serves as an intermediate anchor that provides a reliable geodesic direction, guiding the adversarial feature toward a region that is better aligned with the semantics of its clean counterpart.

A key question is how far the original feature should move along this direction.
Using a fixed step size is suboptimal: a large step may over-correct clean samples, while a small step may be insufficient for adversarial ones.
Since the input may be either clean or adversarial at test time, AGC determines the correction strength adaptively from two complementary signals:
\begin{equation}
\mathrm{dev}
=
\frac{1}{N}
\sum_{i=1}^{N}
\left(1-z_i^\top z\right),
\qquad
\mathrm{rel}
=
\frac{1}{N(N-1)}
\sum_{i=1}^{N}
\sum_{j\neq i}
z_i^\top z_j  .
\end{equation}
Here, $\mathrm{dev}$ measures how far the original feature $z$ deviates from the augmentation-induced neighborhood, while $\mathrm{rel}$ measures the consistency among the augmented views.
Intuitively, a large deviation together with high consistency suggests that the original feature is likely perturbed and that the anchor provides a trustworthy correction direction.

Since $z$ and $z_i$ come from the same image, their similarities $z_i^\top z$ are almost always nonnegative in practice. As a result, although $\mathrm{dev}$ is theoretically bounded in $[0,2]$, it typically lies close to the range $[0,1]$. We therefore clip $\mathrm{dev}$ to $[0,1]$ to avoid excessive sensitivity to rare large deviations.
Meanwhile, $\mathrm{rel}$ is an average pairwise cosine similarity and naturally lies in $[-1,1]$, so we rescale it to $[0,1]$ by $(\mathrm{rel}+1)/2$.
With these normalized signals, the correction score is defined as
\begin{equation}
s_{\mathrm{corr}}
=
\mathrm{dev}^{\gamma}
\cdot
\mathrm{rel},
\end{equation}

where $\gamma$ is a trade-off coefficient. The geodesic step is then defined by interpolating between a conservative step $\beta_{\mathrm{clean}}$ for clean inputs and a stronger step $\beta_{\mathrm{adv}}$ for adversarial inputs:
\begin{equation}
\beta
=
\beta_{\mathrm{clean}}
+
(\beta_{\mathrm{adv}}-\beta_{\mathrm{clean}})\, s_{\mathrm{corr}}.
\end{equation}
Thus, AGC applies a small correction when the input appears clean or the anchor is unreliable, and a larger correction when the input appears adversarial and the augmented views are mutually consistent.

Given the adaptive step coefficient $\beta$, AGC moves the original feature along the geodesic defined by the anchor $a^\star$.
Let
\begin{equation}
u^\star=u(z,a^\star),
\qquad
\theta^\star=\arccos(z^\top a^\star).
\end{equation}
The corrected feature is then given by
\begin{equation}
\tilde{z}
=
\cos(\beta\theta^\star)\,z
+
\sin(\beta\theta^\star)\,u^\star.
\end{equation}
Here, $a^\star$ serves as a reference anchor rather than the final destination of correction. Accordingly, $\beta$ determines whether the update stops before reaching $a^\star$, reaches it exactly, or moves beyond it along the same geodesic.

The final prediction is then obtained by
\begin{equation}
\hat{y}_{\mathrm{final}}
=
\arg\max_c \tilde{z}^{\top} t_c .
\end{equation}

Overall, AGC uses the selected augmentation to define a reliable anchor direction and employs an adaptive step to correct input features.
It requires only forward passes through the frozen CLIP encoders, with no backpropagation or parameter updates at test time, making it an efficient plug-and-play defense.











\section{Experiments}

\subsection{Setup}

\paragraph{Datasets and Models.}
To evaluate the effectiveness of our method, we conduct experiments on eight classification datasets covering diverse visual domains, including general objects (Caltech101~\cite{fei2004learning}), animals (OxfordPets~\cite{parkhi2012cats}), plants (Flower102~\cite{nilsback2008automated}), vehicles (StanfordCars~\cite{krause20133d}, FGVCAircraft~\cite{maji2013fine}), complex textures (DTD~\cite{cimpoi2014describing}), satellite images (EuroSAT~\cite{helber2019eurosat}), and human actions from videos (UCF101~\cite{soomro2012ucf101}). 
For the CLIP backbone~\cite{radford2021learning}, we adopt three standard Vision Transformer architectures: ViT-B/32, ViT-B/16, and ViT-L/14.

\paragraph{Evaluation and Baselines.}
To evaluate adversarial robustness, we follow the evaluation protocol of TTP~\cite{li2025ttp} and report both clean accuracy (\textbf{Acc}) and robust accuracy (\textbf{Rob}) under PGD~\cite{PGD} attacks. 
We compare AGC with three CLIP test-time defense baselines: TTC~\cite{xing2025clip}, R-TPT~\cite{sheng2025r}, and TTP~\cite{li2025ttp}. We also include two test-time adaptation baselines, \textit{Ensemble} and MTA~\cite{zanella2024test}, as well as the original zero-shot predictions of CLIP~\cite{radford2021learning}. 
Here, \textit{Ensemble} averages predictions over multiple views generated by our selected augmentation type.

\paragraph{Implementation Details.}
For adversarial evaluation, we adopt PGD~\cite{PGD} attacks with $\epsilon = 4/255$, $\texttt{steps}=100$. Our selected augmentation type is RandomPerspective with \texttt{distortion\_scale}=0.5. The number of augmented views is set to 32.
For adaptive geodesic correction, we set $\beta_{\mathrm{clean}}=0.45$, $\beta_{\mathrm{adv}}=2.25$ and $\gamma=0.9$. All experiments are conducted on an NVIDIA A100 GPU.

\subsection{Results}

\paragraph{Results on various datasets.}
We first evaluate AGC on eight fine-grained classification datasets. As shown in Table~\ref{tab:results_b32}, with ViT-B/32 as the backbone, AGC achieves an average robust accuracy of \textbf{82.9\%} under 100-step PGD attacks with $\epsilon=4/255$, substantially outperforming all competing methods. In particular, compared with the previous advanced test-time defenses TTP and R-TPT, AGC improves the average robust accuracy by \textbf{43.2\%} and \textbf{47.6\%}, respectively. This demonstrates the effectiveness of performing correction along selected geodesic directions in CLIP's hyperspherical feature space.

A notable finding is that even a simple Ensemble of our selected augmentations outperforms existing advanced defenses. This not only validates the effectiveness of our margin-based augmentation selection criterion but also confirms the strong potential of augmentation features for robustness correction. However, \textit{Ensemble} only combines augmented predictions at the output level, without explicitly leveraging augmented features as semantic anchors to correct adversarial features. Moreover, a simple average cannot fully resolve the trade-off between improving robustness and preserving clean accuracy. In contrast, AGC explicitly uses the selected augmentation feature as a geometric anchor and adaptively corrects the input feature toward it. Compared with \textit{Ensemble}, AGC not only achieves better clean accuracy, but also delivers substantially higher robust accuracy, demonstrating a much more effective way to exploit the robustness-correction capability of augmentations.
\definecolor{lightgray}{gray}{0.9}
\definecolor{highlightred}{RGB}{220, 50, 100}

\begin{table}[!t]
    \centering
    \resizebox{\textwidth}{!}{%
        \begin{tabular}{l *{9}{cc}}
            \toprule
            \multirow{2}{*}{Method}
            & \multicolumn{2}{c}{Caltech101}
            & \multicolumn{2}{c}{Pets}
            & \multicolumn{2}{c}{Cars}
            & \multicolumn{2}{c}{Flower102}
            & \multicolumn{2}{c}{Aircraft}
            & \multicolumn{2}{c}{DTD}
            & \multicolumn{2}{c}{EuroSAT}
            & \multicolumn{2}{c}{UCF101}
            & \multicolumn{2}{c}{Avg.} \\
            & Acc. & Rob.
            & Acc. & Rob.
            & Acc. & Rob.
            & Acc. & Rob.
            & Acc. & Rob.
            & Acc. & Rob.
            & Acc. & Rob.
            & Acc. & Rob.
            & Acc. & Rob. \\
            \midrule
            CLIP
            & 91.4 & \cellcolor{lightgray}0.2
            & 85.1 & \cellcolor{lightgray}0.0
            & 60.1 & \cellcolor{lightgray}0.0
            & 64.0 & \cellcolor{lightgray}0.0
            & 18.1 & \cellcolor{lightgray}0.0
            & 43.0 & \cellcolor{lightgray}0.0
            & 35.8 & \cellcolor{lightgray}0.0
            & 61.6 & \cellcolor{lightgray}0.0
            & 57.4 & 0.0 \\

            TTC
            & 86.5 & 22.7
            & 83.5 & 11.8
            & 48.1 & 2.3
            & 64.3 & 3.2
            & 18.2 & 1.0
            & 37.3 & 4.7
            & \textbf{53.0} & 3.0
            & 62.6 & 6.1
            & 56.7 & 6.9 \\

            Ensemble
            & 91.7 & 83.8
            & 82.3 & 66.2
            & 59.8 & 33.1
            & 61.7 & 43.2
            & 19.5 & 9.3
            & 39.6 & 30.1
            & 26.3 & 10.1
            & 59.3 & 44.1
            & 55.9 & 40.0 \\

            MTA
            & 92.0 & 76.3
            & \textbf{86.3} & 53.6
            & \textbf{63.4} & 26.4
            & \textbf{64.4} & 36.5
            & \textbf{20.2} & 8.2
            & \textbf{43.8} & 28.8
            & 34.6 & 11.3
            & \textbf{63.3} & 39.1
            & \textbf{58.5} & 35.0 \\

            R-TPT
            & 90.6 & 76.4
            & 84.5 & 55.8
            & 63.1 & 28.4
            & 62.6 & 37.6
            & 19.1 & 9.2
            & 42.1 & 29.1
            & 32.0 & 5.1
            & 62.8 & 41.0
            & 57.1 & 35.3 \\

            TTP
            & 90.9 & 81.8
            & 84.7 & 61.0
            & 59.8 & 29.8
            & 63.6 & 42.0
            & 18.0 & 10.3
            & 42.8 & 32.2
            & 35.6 & 14.1
            & 61.3 & 46.6
            & 57.1 & 39.7 \\

            AGC (Ours)
            & \textbf{92.2} & \textcolor{highlightred}{\textbf{97.9}}
            & 84.1 & \textcolor{highlightred}{\textbf{94.0}}
            & 62.2 & \textcolor{highlightred}{\textbf{87.4}}
            & 63.1 & \textcolor{highlightred}{\textbf{71.5}}
            & 19.9 & \textcolor{highlightred}{\textbf{56.1}}
            & 41.2 & \textcolor{highlightred}{\textbf{71.2}}
            & 30.0 & \textcolor{highlightred}{\textbf{93.4}}
            & 60.5 & \textcolor{highlightred}{\textbf{91.3}}
            & 56.7 & \textcolor{highlightred}{\textbf{82.9}} \\
            \bottomrule
        \end{tabular}
    }
    \caption{Adversarial (Rob.) and Clean (Acc.) accuracy (\%) on fine-grained classification datasets with pre-trained ViT-B/32 ($\epsilon = 4 / 255$).}
    \label{tab:results_b32}
\end{table}
\begin{table}[!t]
    \centering
    \resizebox{\textwidth}{!}{%
        \begin{tabular}{l *{9}{cc}}
            \toprule
            \multirow{2}{*}{Method}
            & \multicolumn{2}{c}{Caltech101}
            & \multicolumn{2}{c}{Pets}
            & \multicolumn{2}{c}{Cars}
            & \multicolumn{2}{c}{Flower102}
            & \multicolumn{2}{c}{Aircraft}
            & \multicolumn{2}{c}{DTD}
            & \multicolumn{2}{c}{EuroSAT}
            & \multicolumn{2}{c}{UCF101}
            & \multicolumn{2}{c}{Avg.} \\
            & Acc. & Rob.
            & Acc. & Rob.
            & Acc. & Rob.
            & Acc. & Rob.
            & Acc. & Rob.
            & Acc. & Rob.
            & Acc. & Rob.
            & Acc. & Rob.
            & Acc. & Rob. \\
            \midrule
            CLIP
            & 94.0 & \cellcolor{lightgray}0.0
            & \textbf{88.3} & \cellcolor{lightgray}0.0
            & 65.5 & \cellcolor{lightgray}0.0
            & 67.4 & \cellcolor{lightgray}0.0
            & 23.9 & \cellcolor{lightgray}0.0
            & 44.4 & \cellcolor{lightgray}0.0
            & 42.2 & \cellcolor{lightgray}0.0
            & 65.2 & \cellcolor{lightgray}0.0
            & 61.4 & 0.0 \\

            TTC
            & 87.6 & 8.4
            & 82.3 & 10.4
            & 55.0 & 2.9
            & \textbf{69.0} & 7.4
            & 23.3 & 0.5
            & 41.0 & 4.5
            & \textbf{47.4} & 0.4
            & 65.8 & 1.6
            & 58.9 & 4.5 \\

            Ensemble
            & 94.2 & 84.4
            & 85.0 & 65.9
            & 66.0 & 41.8
            & 67.2 & 46.9
            & 23.5 & 12.7
            & 41.7 & 31.9
            & 32.7 & 14.8
            & 63.9 & 43.4
            & 59.3 & 42.7 \\

            MTA
            & 94.3 & 72.1
            & 88.0 & 51.8
            & \textbf{67.7} & 18.5
            & 67.4 & 27.9
            & \textbf{25.0} & 4.3
            & \textbf{46.5} & 16.2
            & 42.5 & 1.2
            & \textbf{67.5} & 27.5
            & \textbf{62.4} & 27.4 \\

            R-TPT
            & 93.7 & 82.0
            & 87.2 & 60.2
            & 67.0 & 34.7
            & 68.7 & 44.6
            & 23.9 & 13.2
            & 46.4 & 32.8
            & 34.7 & 8.5
            & 67.2 & 43.2
            & 61.1 & 39.9 \\

            TTP
            & 93.5 & 82.3
            & \textbf{88.3} & 64.7
            & 65.4 & 37.4
            & 67.3 & 47.2
            & 23.9 & 14.8
            & 44.1 & 36.0
            & 42.0 & 14.5
            & 65.0 & 47.2
            & 61.2 & 43.0 \\

            AGC (Ours)
            & \textbf{94.4} & \textcolor{highlightred}{\textbf{99.4}}
            & 87.0 & \textcolor{highlightred}{\textbf{96.1}}
            & 67.3 & \textcolor{highlightred}{\textbf{93.2}}
            & 67.8 & \textcolor{highlightred}{\textbf{89.6}}
            & 24.2 & \textcolor{highlightred}{\textbf{68.6}}
            & 43.4 & \textcolor{highlightred}{\textbf{88.4}}
            & 36.0 & \textcolor{highlightred}{\textbf{97.7}}
            & 65.2 & \textcolor{highlightred}{\textbf{98.0}}
            & 60.7 & \textcolor{highlightred}{\textbf{91.4}} \\
            \bottomrule
        \end{tabular}
    }
    \caption{Adversarial (Rob.) and Clean (Acc.) accuracy (\%) on fine-grained classification datasets with pre-trained ViT-B/16 ($\epsilon = 4 / 255$).}
    \label{tab:results_b16}
\end{table}
\paragraph{Results on different backbones.}
We further evaluate AGC on two different CLIP backbones, ViT-B/16 and ViT-L/14, and report the results in Tables~\ref{tab:results_b16} and~\ref{tab:results_l14}. AGC continues to show clear robustness advantages on both architectures. Specifically, it achieves average robust accuracies of \textbf{91.4\%} on ViT-B/16 and \textbf{93.4\%} on ViT-L/14, far exceeding the previous best results of \textbf{43.0\%} and \textbf{51.6\%}, respectively. We also include results on ResNet-based backbones in Appendix~\ref{ResNet}.

Meanwhile, AGC achieves clean accuracy comparable to vanilla CLIP. It shows that AGC consistently improves robustness across different CLIP backbones while preserving their original zero-shot generalization ability. It further suggests that both the margin criterion and the adaptive step size strategy generalize well across backbones. Since AGC requires no parameter updates, it serves as a truly plug-and-play defense for vision-language models.

\begin{table}[!t]
    \centering
    \resizebox{\textwidth}{!}{%
        \begin{tabular}{l *{9}{cc}}
            \toprule
            \multirow{2}{*}{Method}
            & \multicolumn{2}{c}{Caltech101}
            & \multicolumn{2}{c}{Pets}
            & \multicolumn{2}{c}{Cars}
            & \multicolumn{2}{c}{Flower102}
            & \multicolumn{2}{c}{Aircraft}
            & \multicolumn{2}{c}{DTD}
            & \multicolumn{2}{c}{EuroSAT}
            & \multicolumn{2}{c}{UCF101}
            & \multicolumn{2}{c}{Avg.} \\
            & Acc. & Rob.
            & Acc. & Rob.
            & Acc. & Rob.
            & Acc. & Rob.
            & Acc. & Rob.
            & Acc. & Rob.
            & Acc. & Rob.
            & Acc. & Rob.
            & Acc. & Rob. \\
            \midrule
            CLIP
            & 95.2 & \cellcolor{lightgray}0.1
            & 93.1 & \cellcolor{lightgray}0.0
            & 76.8 & \cellcolor{lightgray}0.0
            & 76.2 & \cellcolor{lightgray}0.0
            & 30.0 & \cellcolor{lightgray}0.0
            & 52.4 & \cellcolor{lightgray}0.0
            & 55.1 & \cellcolor{lightgray}0.0
            & 73.7 & \cellcolor{lightgray}0.0
            & \textbf{69.1} & 0.0 \\

            TTC
            & 88.7 & 7.7
            & 92.2 & 7.6
            & 67.8 & 2.2
            & \textbf{76.5} & 7.5
            & 31.7 & 0.5
            & 49.7 & 6.2
            & \textbf{64.1} & 0.2
            & \textbf{75.0} & 2.2
            & 68.2 & 4.3 \\

            Ensemble
            & 95.3 & 89.7
            & 92.5 & 77.6
            & 76.6 & 54.4
            & 75.0 & 60.0
            & 29.4 & 17.8
            & 50.2 & 38.7
            & 49.8 & 27.8
            & 72.6 & 57.3
            & 67.7 & 52.9 \\

            MTA
            & \textbf{95.8} & 83.1
            & \textbf{93.7} & 64.9
            & \textbf{78.4} & 36.6
            & 76.1 & 44.2
            & \textbf{32.7} & 8.0
            & 53.4 & 27.2
            & 47.8 & 7.5
            & 74.7 & 47.5
            & \textbf{69.1} & 39.9 \\

            R-TPT
            & 95.7 & 88.2
            & \textbf{93.7} & 72.9
            & 77.2 & 49.1
            & 76.2 & 55.6
            & 31.7 & 17.2
            & \textbf{54.0} & 38.0
            & 44.3 & 20.4
            & 74.3 & 55.6
            & 68.4 & 49.6 \\

            TTP
            & 95.1 & 88.6
            & 93.1 & 76.3
            & 76.8 & 51.1
            & 76.1 & 58.7
            & 29.2 & 17.7
            & 52.3 & 41.3
            & 55.0 & 21.6
            & 73.6 & 57.4
            & 68.9 & 51.6 \\

            AGC (Ours)
            & 95.6 & \textcolor{highlightred}{\textbf{97.9}}
            & 93.4 & \textcolor{highlightred}{\textbf{98.5}}
            & 77.3 & \textcolor{highlightred}{\textbf{95.4}}
            & 76.0 & \textcolor{highlightred}{\textbf{96.0}}
            & 30.5 & \textcolor{highlightred}{\textbf{77.7}}
            & 52.2 & \textcolor{highlightred}{\textbf{95.2}}
            & 52.0 & \textcolor{highlightred}{\textbf{89.0}}
            & 73.1 & \textcolor{highlightred}{\textbf{97.4}}
            & 68.8 & \textcolor{highlightred}{\textbf{93.4}} \\
            \bottomrule
        \end{tabular}
    }
    \caption{Adversarial (Rob.) and Clean (Acc.) accuracy (\%) on fine-grained classification datasets with pre-trained ViT-L/14 ($\epsilon = 4 / 255$).}
    \label{tab:results_l14}
\end{table}

\begin{wraptable}{r}{0.45\textwidth} 
    \centering
    \scriptsize
    \vspace{-5pt}
    \begin{tabular}{lcccc}
        \hline
        Input & Vanilla & R-TPT & TTP & AGC \\
        \hline
        Clean & 0.0057s & 0.27s & 0.0060s & 0.0091s \\
        Adv. & 0.0057s & 0.27s & 0.097s & 0.0091s \\
        \hline
    \end{tabular}
    \caption{Average inference time per sample on an A100 GPU.}
    \label{tab:inference_time}
\end{wraptable}
\paragraph{Defense efficiency.}
Another important advantage of AGC over prior test-time defenses is that it completely avoids gradient backpropagation and parameter updates during inference, which substantially reduces computational overhead. To verify this practical benefit, we report the average inference time per sample on an A100 GPU in Table~\ref{tab:inference_time}. AGC takes 0.0091s on input samples. On adversarial inputs, AGC is about 10$\times$ faster than TTP, since it relies only on forward feature extraction rather than backpropagation-based test-time optimization. TTP is slightly faster on clean samples because it first performs a clean/adversarial detection step and can bypass expensive adaptation when the input is identified as clean. In contrast, AGC applies the same lightweight forward procedure to all inputs, without any input-type detection. Even so, on clean samples it is still more than 29$\times$ faster than R-TPT, which likewise performs no such pre-detection. Combined with its substantially stronger robustness, these results make AGC a more practical defense.


\subsection{More Analysis}


\paragraph{Augmentation selection.}
Our earlier analysis identifies \textit{RandomPerspective} as the strongest single augmentation type. A natural follow-up question is whether robustness can be further improved by combining it with more augmentation types. To investigate this, we study how augmentation-type selection affects robustness on the Caltech101 dataset. We consider six best-performing augmentation types: \textit{RandomPerspective}, \textit{Random\_Crop}, \textit{center\_crop}, \textit{Rotation}, \textit{Random\_Resized\_crop}, and \textit{hflip}. To ensure a fair comparison as the number of augmentation types increases, we fix the total number of augmented views to 64.

As shown in Fig.~\ref{fig:aug_type}, adding more augmentation types leads to a clear drop in robust accuracy, while clean accuracy remains relatively stable. This suggests that robustness does not improve simply by increasing augmentation diversity. Instead, mixing different augmentation types likely introduces inconsistent directions in the feature space, which weakens the semantic correction signal. Therefore, we use only \textit{RandomPerspective} as the default augmentation type in AGC. Due to space limitations, we defer the ablation study of the distortion in \textit{RandomPerspective} to Appendix~\ref{distortion}.



\paragraph{Number of views.}
After selecting \textit{RandomPerspective} as the default augmentation type, we further study the effect of the number of augmented views. As shown in Fig.~\ref{fig:num_views}, both clean and robust accuracy remain relatively stable when the number of views increases, with only minor fluctuations. This indicates that the robustness gain of AGC mainly comes from selecting a semantically reliable augmentation type, rather than from using a large number of augmented views for weighting. Once the augmentation type is well chosen, only a small number of views is sufficient to estimate a stable geometric anchor for correction. This further demonstrates the efficiency of AGC.





\begin{wrapfigure}{r}{0.5\textwidth}
    \centering
     \vspace{-12pt}
    \begin{subfigure}[t]{0.24\textwidth}
        \centering
        \includegraphics[width=\linewidth]{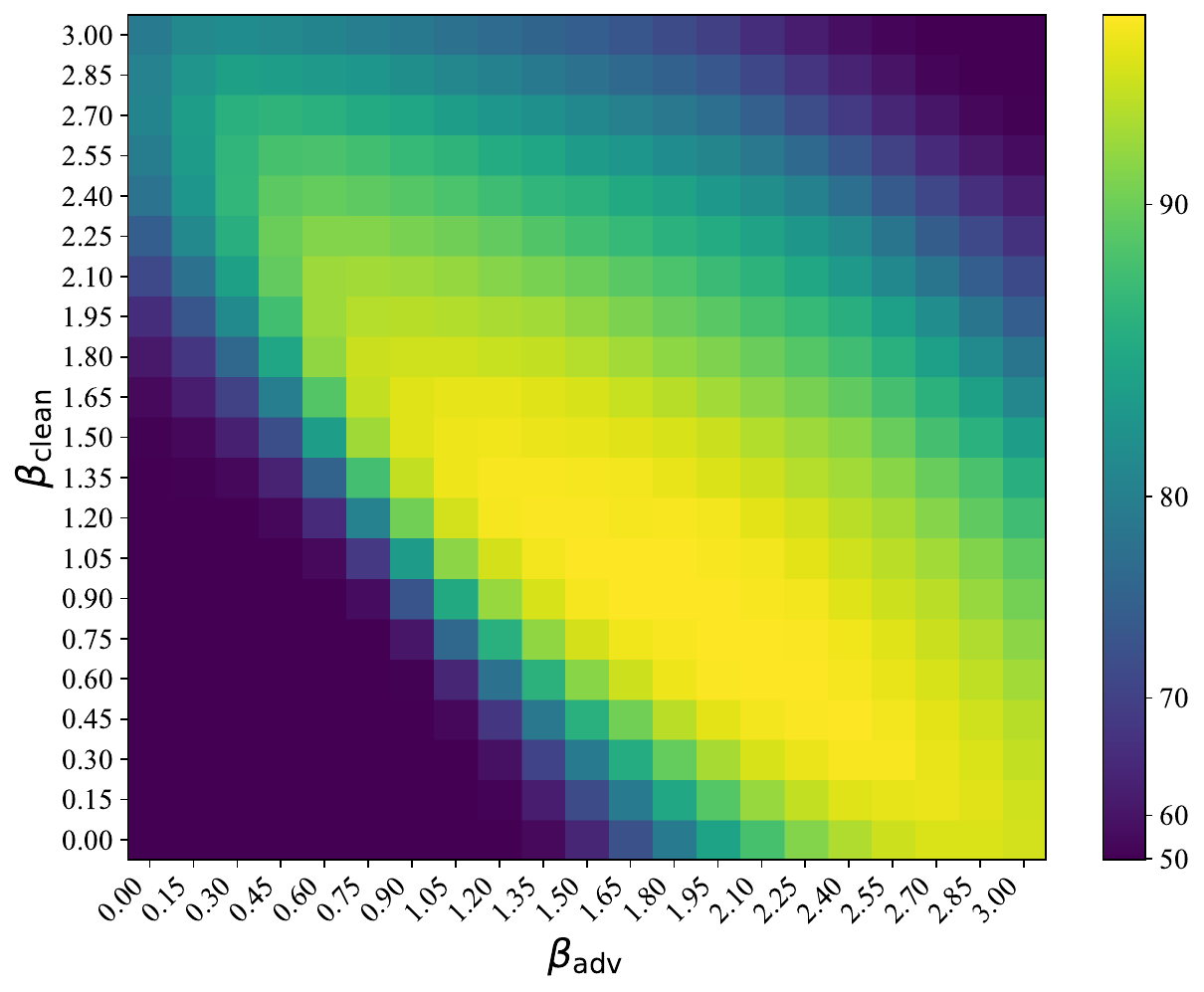}
    \end{subfigure}
    \hfill
    \begin{subfigure}[t]{0.24\textwidth}
        \centering
        \includegraphics[width=\linewidth]{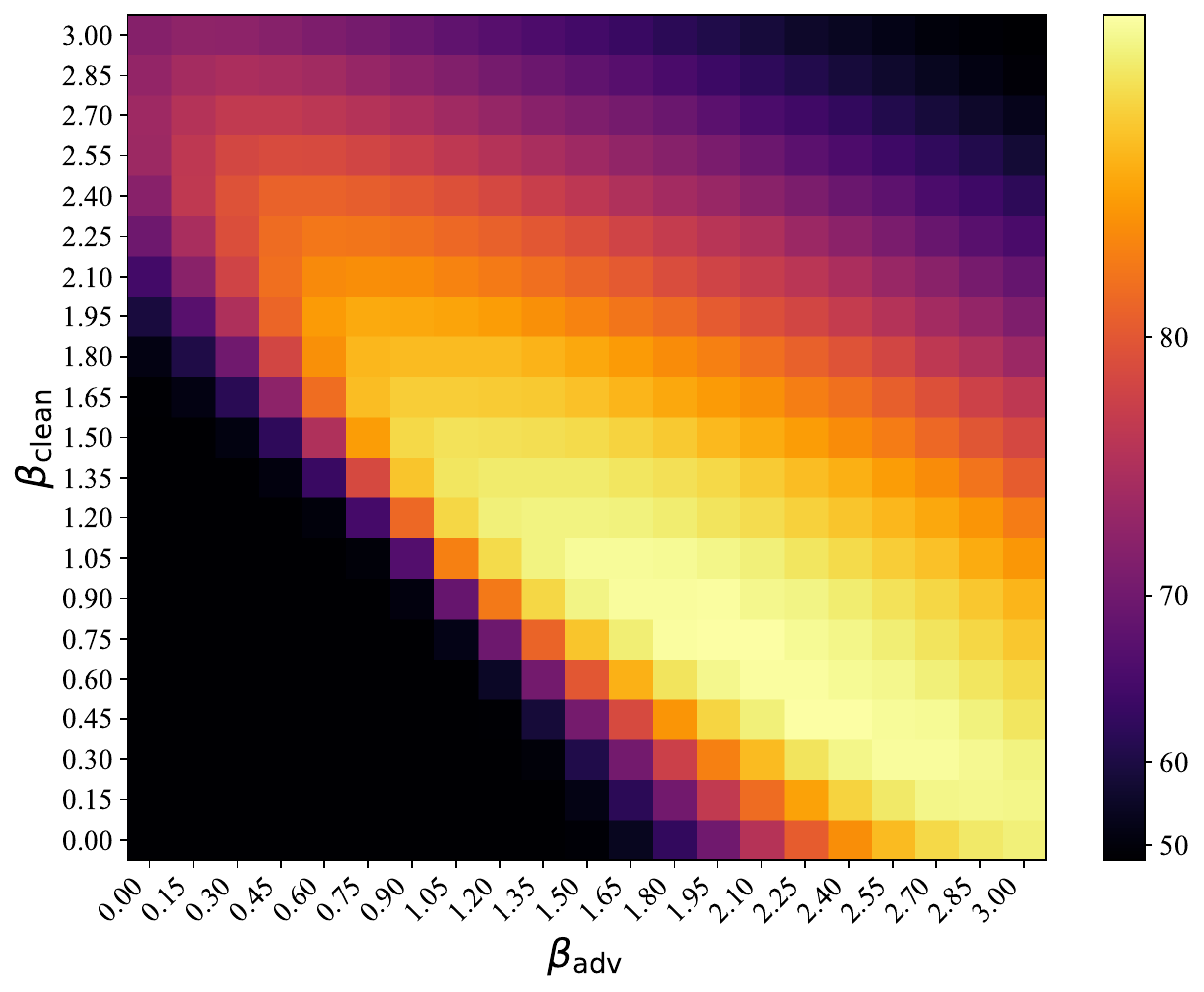}
    \end{subfigure}
    \caption{Mean Acc. on Caltech101 and Pets Using different $\beta$.}
    \label{fig:beta_ablation}
\end{wrapfigure}

\paragraph{Parameter $\beta$ settings.}
In AGC, the adaptive step size is mainly controlled by the hyperparameters $\beta_{\mathrm{clean}}$ and $\beta_{\mathrm{adv}}$. To evaluate their joint effect, we perform a grid search over $[0, 3]$ with an interval of 0.15, measuring the average of clean and robust accuracy ($(\mathrm{Acc}+\mathrm{Rob})/2$) on Caltech101, Pets and UCF101. Fig.~\ref{fig:beta_ablation} reveals a highly consistent performance landscape across both datasets, indicating strong cross-domain generalization. The peak performance falls squarely within the region where $\beta_{\mathrm{clean}}\in[0.3,0.6]$ and $\beta_{\mathrm{adv}}\in[2.1,2.6]$. Accordingly, we adopt $\beta_{\mathrm{clean}}=0.45$ and $\beta_{\mathrm{adv}}=2.25$ as the default settings in all our experiments. Due to space limitations, we defer the ablation study of the trade-off coefficient $\gamma$ to Appendix~\ref{app:gamma}.

\begin{minipage}[!t]{0.31\textwidth}
    \centering
    \includegraphics[width=\linewidth]{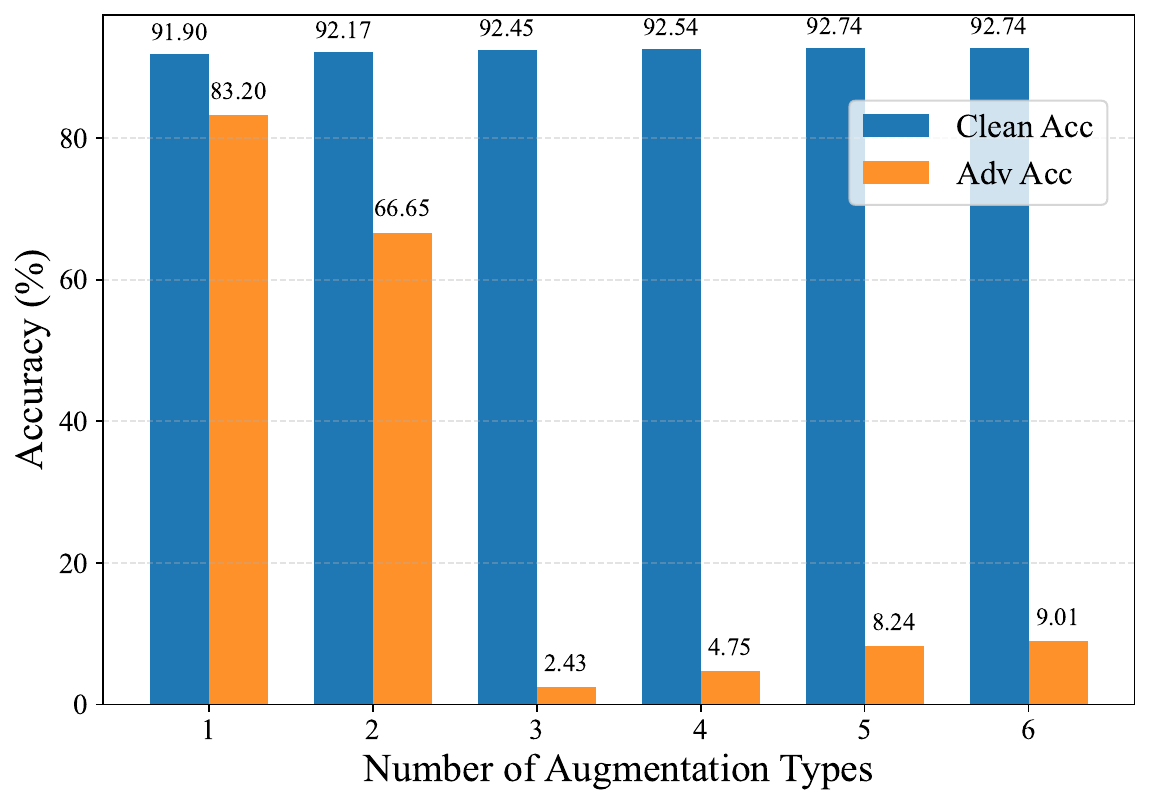}
    \captionof{figure}{Effect of augmentation type selection on clean and robust accuracy.}
    \label{fig:aug_type}
\end{minipage}
\hfill
\begin{minipage}[!t]{0.31\textwidth}
    \centering
    \includegraphics[width=\linewidth]{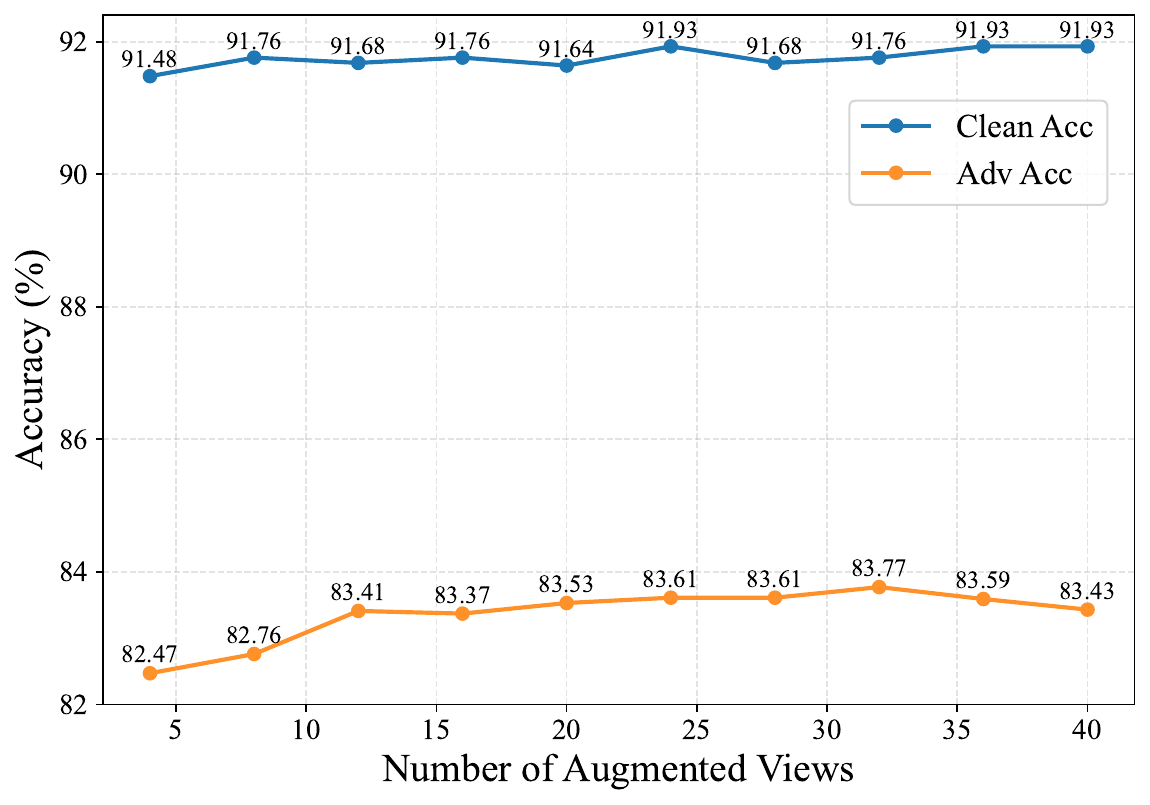}
    \captionof{figure}{Effect of the number of augmented views on clean and robust accuracy.}
    \label{fig:num_views}
\end{minipage}
\hfill
\begin{minipage}[!t]{0.31\textwidth}
    \centering
    \includegraphics[width=\linewidth]{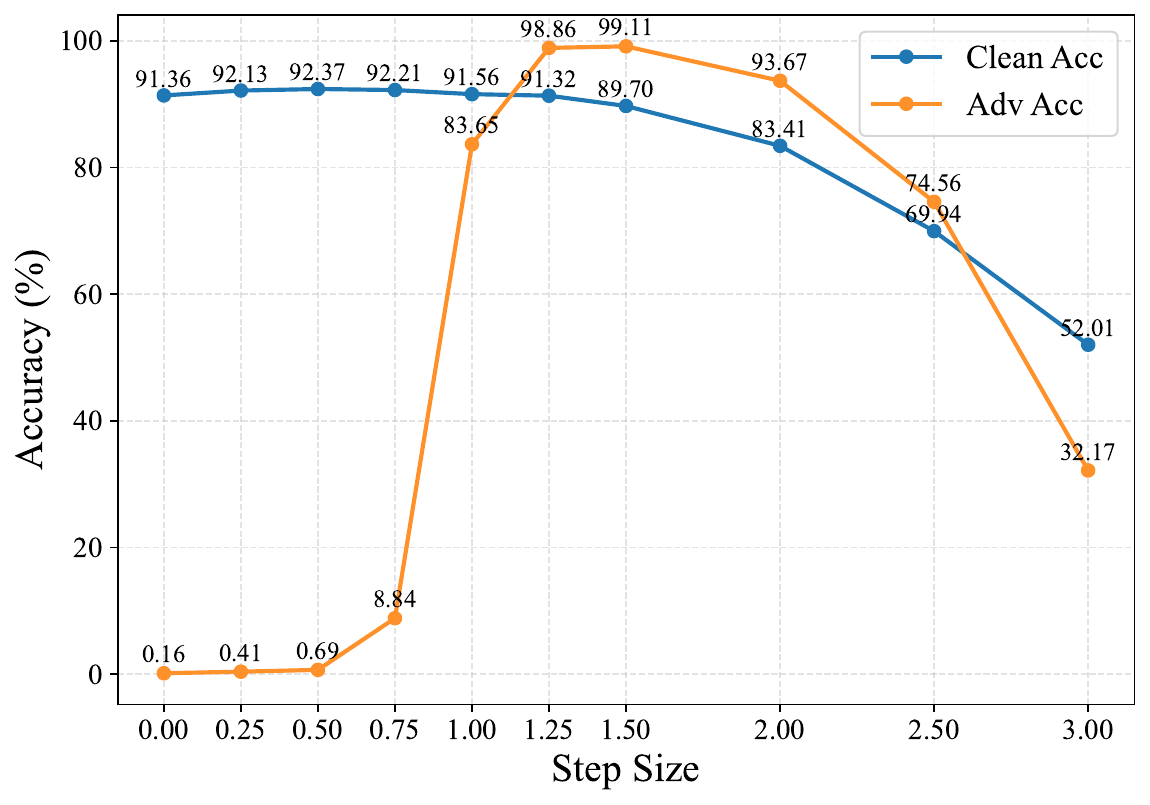}
    \captionof{figure}{Effect of fixed step scale on clean and robust accuracy.}
    \label{fig:acc_step_scale}
\end{minipage}

\paragraph{Step scale.}
We next examine whether a fixed correction step is sufficient or whether adaptive step sizing is necessary. To answer this, we evaluate AGC on Caltech101 with a range of fixed step scales. As shown in Fig.~\ref{fig:acc_step_scale}, the optimal clean accuracy and the optimal robust accuracy occur at different step scales, meaning that no single fixed step can simultaneously achieve the best of both. This directly motivates the need for an adaptive mechanism.

Under fixed scales, the peak clean and robust accuracies reach \textbf{92.37\%} and \textbf{99.11\%}, respectively. By comparison, as reported in Table~\ref{tab:results_b32}, adaptive AGC achieves \textbf{92.2\%} clean accuracy and \textbf{97.9\%} robust accuracy on the same dataset. These results show that AGC can automatically strike a strong balance between preserving clean semantics and correcting adversarial features, without manual tuning of the step size.


\section{Discussion}
AGC suggests a broader perspective on test-time defense for vision-language models. In our experiments, simply ensembling the strongest selected augmentation already surpasses prior state-of-the-art defenses, indicating that CLIP itself has a strong self-correction ability under suitable transformations. While prior work such as TTC~\cite{xing2025clip} touches on this phenomenon, AGC shows for the first time that it can be exploited effectively without any parameter updates at test time. Although AGC’s hyperparameters generalize well across diverse datasets in our experiments, the method still relies on several predefined ones. Developing a fully hyperparameter-free, update-free defense is therefore an important direction for our future research.

\section{Conclusion}

In this paper, we present AGC, a truly training-free test-time defense for CLIP, motivated by the observation that augmentations are not equally robust. By analyzing augmented adversarial features in CLIP’s normalized feature space, we show that robust augmentations induce geodesic directions that better align with clean class semantics, which makes them reliable anchors for feature correction. Building on this insight, AGC adaptively corrects features using only the best augmentation, without any gradient-based optimization or parameter updates at inference. Extensive experiments across eight fine-grained datasets and three CLIP backbones show that AGC substantially outperforms prior test-time defenses in adversarial robustness while remaining highly efficient and preserving clean accuracy. These findings highlight the value of understanding augmentation through geometry and suggest a promising direction for developing practical robustness methods for vision-language models.

\bibliographystyle{unsrt}
\bibliography{references}

\newpage
\appendix

\section{Overview}
This appendix presents detailed dataset information and additional experiments omitted from the main paper due to space constraints.

\section{Datasets}
As shown in Table~\ref{tab:datasets}, we present the number of categories and images of all datasets used in our experiments, following the same dataset setup as TTP~\cite{li2025ttp} and R-TPT~\cite{sheng2025r}.
\begin{table}[htbp]
    \centering
    \begin{tabular}{l l c r}
        \toprule
        \textbf{Dataset} & \textbf{Description} & \textbf{\# Classes} & \textbf{\# Test} \\
        \midrule
        Caltech101 & Object images & 100 & 2,465 \\
        Pets & Pet images & 37 & 3,669 \\
        Cars & Car images & 196 & 8,041 \\
        Flower102 & Flower images & 102 & 2,463 \\
        Aircraft & Aircraft images & 100 & 3,333 \\
        DTD & Describable textures dataset & 47 & 1,692 \\
        EuroSAT & Sentinel-2 satellite images & 10 & 8,100 \\
        UCF101 & Human action images & 101 & 3,783 \\
        \bottomrule
    \end{tabular}
    \caption{All datasets involved in our experiments.}
    \label{tab:datasets}
\end{table}

\section{Results of Different $\gamma$}
\label{app:gamma}
\begin{figure}[htbp]
    \centering

    \begin{subfigure}[t]{0.45\linewidth}
        \centering
        \includegraphics[width=\linewidth]{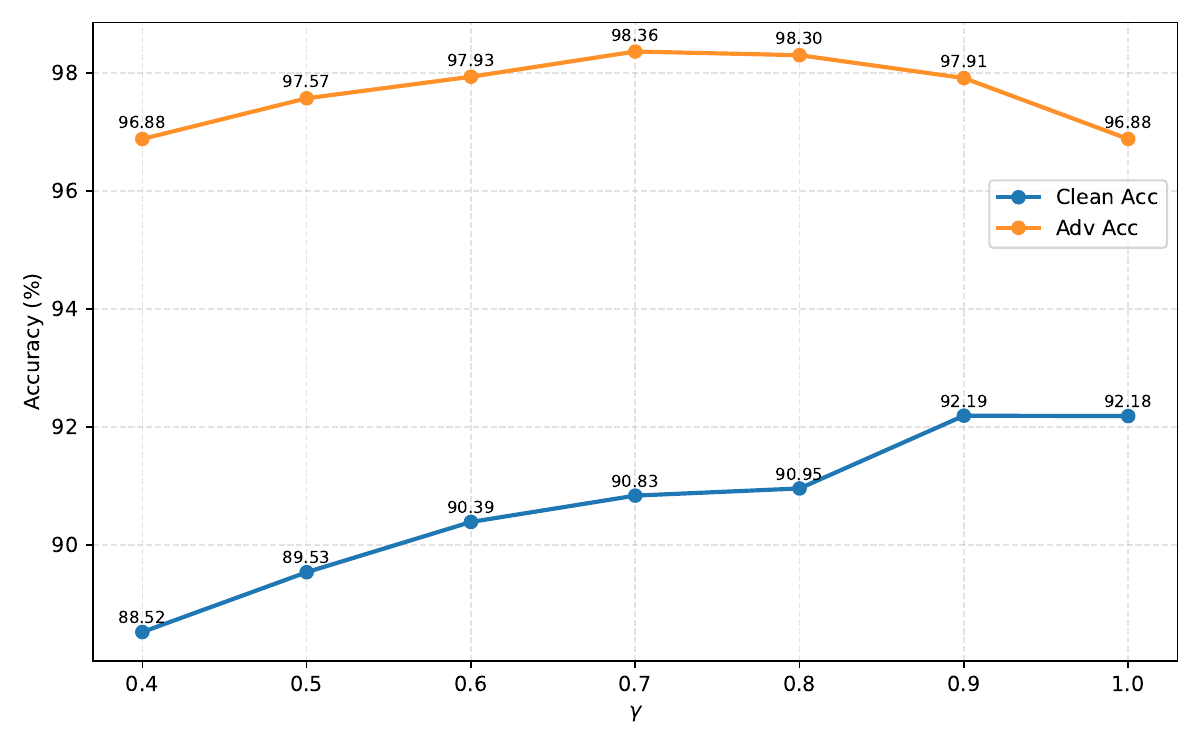}
        \caption{Accuracy under different $\gamma$ values}
        \label{fig:acc_gamma}
    \end{subfigure}
    \hfill
    \begin{subfigure}[t]{0.45\linewidth}
        \centering
        \includegraphics[width=\linewidth]{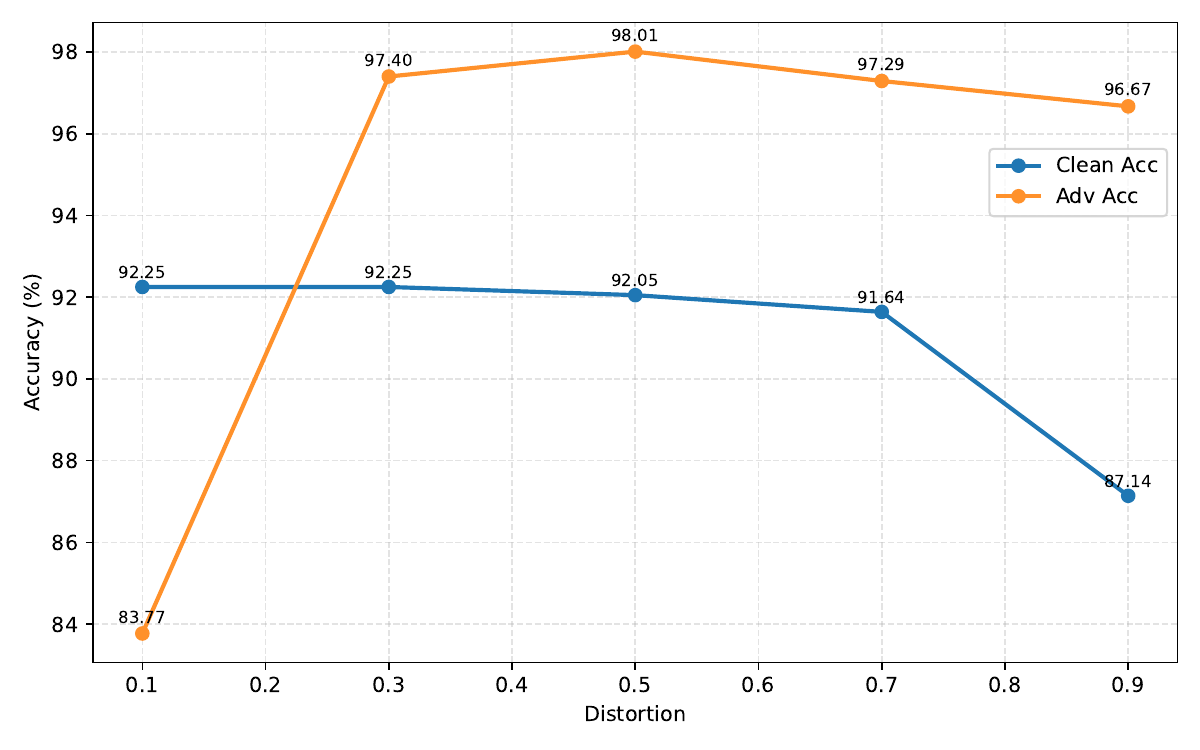}
        \caption{Accuracy under different distortion scales}
        \label{fig:acc_distortion}
    \end{subfigure}

    \caption{Clean and adversarial accuracy under different hyperparameter settings.}
    \label{fig:acc_hyperparams}
\end{figure}

For completeness, we further analyze the effect of the hyperparameter $\gamma$ using CLIP ViT-B/32 on Caltech101. As shown in Figure~\ref{fig:acc_gamma}, the accuracy on adversarial examples initially increases and then decreases, peaking at around $\gamma=0.7$. Meanwhile, the clean accuracy steadily improves with larger values of $\gamma$ and reaches its maximum at around $\gamma=0.9$. Although $\gamma=0.9$ leads to slightly lower adversarial accuracy than $\gamma=0.7$ or $\gamma=0.8$, it provides a better trade-off between clean and adversarial performance, resulting in the highest average accuracy. Therefore, we set $\gamma=0.9$ as the default value.

\section{Results under different Distortion in \textit{RandomPerspective}}
\label{distortion}
AGC selects \textit{RandomPerspective} as the anchor augmentation. However, \textit{RandomPerspective} itself contains a distortion parameter that controls the transformation strength. For a more rigorous evaluation, we further conduct an ablation study on this distortion scale using CLIP ViT-B/32 on Caltech101. The results are shown in Figure~\ref{fig:acc_distortion}.  The observed trend is consistent with intuition. As the distortion scale increases, the clean accuracy gradually decreases because stronger distortion increasingly disrupts the original image semantics. In contrast, adversarial robustness first improves as moderate distortion helps correct adversarial features, but begins to decline once the distortion becomes too strong and severely damages the image content. Based on this trade-off, we choose a moderate distortion scale of $0.5$ for all experiments, as it achieves the best adversarial robustness while keeping the drop in clean accuracy limited.

\section{Results under ResNet Backbones}
\label{ResNet}
Due to the limitations of the space, we only present our results on Vision Transformer(ViT) backbones to illustrate our generalizability. To present a more comprehensive evaluation, we conduct an additional experiment on ResNet backbone, which is also a widely used vision module. The result presented in Table~\ref{tab:results_resnet50} demonstrates that our method consistently outperforms all baselines across datasets under the PGD attack.
\begin{table}[htbp]
    \centering
    \resizebox{\textwidth}{!}{%
        \begin{tabular}{l *{9}{cc}}
            \toprule
            \multirow{2}{*}{Method}
            & \multicolumn{2}{c}{Caltech101}
            & \multicolumn{2}{c}{Pets}
            & \multicolumn{2}{c}{Cars}
            & \multicolumn{2}{c}{Flower102}
            & \multicolumn{2}{c}{Aircraft}
            & \multicolumn{2}{c}{DTD}
            & \multicolumn{2}{c}{EuroSAT}
            & \multicolumn{2}{c}{UCF101}
            & \multicolumn{2}{c}{Avg.} \\
            & Acc. & Rob.
            & Acc. & Rob.
            & Acc. & Rob.
            & Acc. & Rob.
            & Acc. & Rob.
            & Acc. & Rob.
            & Acc. & Rob.
            & Acc. & Rob.
            & Acc. & Rob. \\
            \midrule

            CLIP
            & 85.9 & 2.6
            & 83.6 & 0.0
            & 55.7 & 0.0
            & 61.7 & 0.0
            & 15.7 & 0.0
            & 40.4 & 0.8
            & 23.7 & 0.0
            & 59.0 & 0.0
            & 53.2 & 0.4 \\

            TeCoA$^{1}$
            & 78.3 & 78.3
            & 76.0 & 75.8
            & 22.4 & 22.3
            & 33.5 & 33.4
            & 5.8 & 5.8
            & 26.2 & 26.0
            & 16.5 & 16.6
            & 38.4 & 38.1
            & 37.1 & 37.0 \\

            APT$^{1}$
            & 2.9 & 1.7
            & 31.9 & 3.8
            & 8.5 & 0.6
            & 2.6 & 1.1
            & 0.9 & 0.9
            & 16.6 & 7.9
            & 17.0 & 4.0
            & 11.2 & 0.9
            & 11.4 & 2.6 \\

            APT$^{1}$+TeCoA$^{1}$
            & 82.8 & 82.8
            & 79.3 & 79.0
            & 33.9 & 33.6
            & 42.7 & 42.6
            & 9.9 & 9.7
            & 39.2 & 39.0
            & \textbf{32.9} & 32.9
            & 51.5 & 51.4
            & 46.5 & \textbf{46.4} \\

            R-TPT
            & 86.7 & 79.8
            & \textbf{84.6} & 74.2
            & \textbf{58.1} & 42.9
            & 60.6 & 51.9
            & \textbf{17.5} & 12.6
            & \textbf{41.3} & 33.5
            & 21.2 & 15.9
            & \textbf{59.7} & 50.9
            & \textbf{53.7} & 45.2 \\

            TTP
            & 86.0 & 86.4
            & 83.5 & 70.3
            & 55.6 & 41.5
            & \textbf{61.8} & 49.0
            & 16.1 & 11.7
            & 41.0 & 33.3
            & 21.3 & 26.9
            & 59.1 & 50.3
            & 53.0 & 46.2 \\

            AGC(ours)
            & \textbf{88.3} & \textcolor{highlightred}{\textbf{94.9}}
            & 84.0 & \textcolor{highlightred}{\textbf{85.1}}
            & 57.4 & \textcolor{highlightred}{\textbf{80.1}}
            & 61.6 & \textcolor{highlightred}{\textbf{62.8}}
            & 16.6 & \textcolor{highlightred}{\textbf{36.3}}
            & 39.1 & \textcolor{highlightred}{\textbf{53.4}}
            & 21.4 & \textcolor{highlightred}{\textbf{72.7}}
            & 59.4 & \textcolor{highlightred}{\textbf{79.7}}
            & 53.5 & \textcolor{highlightred}{\textbf{70.6}} \\

            \bottomrule
        \end{tabular}
    }
    \caption{Results (\%) of training-time defense methods on fine-grained classification datasets with pre-trained ResNet50 ($\epsilon = 1.0$).}
    \label{tab:results_resnet50}
\end{table}








\end{document}